# Learning Document-Level Semantic Properties
# from Free-Text Annotations

**S.R.K. Branavan**                                    BRANAVAN@CSAIL.MIT.EDU
**Harr Chen**                                              HARR@CSAIL.MIT.EDU
**Jacob Eisenstein**                                   JACOBE@CSAIL.MIT.EDU
**Regina Barzilay**                                    REGINA@CSAIL.MIT.EDU
*Computer Science and Artificial Intelligence Laboratory*
*Massachusetts Institute of Technology*
*77 Massachusetts Avenue, Cambridge MA 02139*

## Abstract

This paper presents a new method for inferring the semantic properties of documents by leveraging free-text keyphrase annotations. Such annotations are becoming increasingly abundant due to the recent dramatic growth in semi-structured, user-generated online content. One especially relevant domain is product reviews, which are often annotated by their authors with pros/cons keyphrases such as "a real bargain" or "good value." These annotations are representative of the underlying semantic properties; however, unlike expert annotations, they are noisy: lay authors may use different labels to denote the same property, and some labels may be missing. To learn using such noisy annotations, we find a hidden paraphrase structure which clusters the keyphrases. The paraphrase structure is linked with a latent topic model of the review texts, enabling the system to predict the properties of unannotated documents and to effectively aggregate the semantic properties of multiple reviews. Our approach is implemented as a hierarchical Bayesian model with joint inference. We find that joint inference increases the robustness of the keyphrase clustering and encourages the latent topics to correlate with semantically meaningful properties. Multiple evaluations demonstrate that our model substantially outperforms alternative approaches for summarizing single and multiple documents into a set of semantically salient keyphrases.

## 1. Introduction

Identifying the document-level semantic properties implied by a text is a core problem in natural language understanding. For example, given the text of a restaurant review, it would be useful to extract a semantic-level characterization of the author's reaction to specific aspects of the restaurant, such as food and service quality (see Figure 1). Learning-based approaches have dramatically increased the scope and robustness of such semantic processing, but they are typically dependent on large expert-annotated datasets, which are costly to produce (Zaenen, 2006).

We propose to use an alternative source of annotations for learning: free-text keyphrases produced by novice users. As an example, consider the lists of pros and cons that often accompany reviews of products and services. Such end-user annotations are increasingly prevalent online, and they grow organically to keep pace with subjects of interest and socio-cultural trends. Beyond such pragmatic considerations, free-text annotations are appealing from a linguistic standpoint because they capture the intuitive semantic judgments of non-specialist language users. In many real-world datasets, these annotations are created by the document's original author, providing a direct window into the semantic judgments that motivated the document text.

| |
|---|
| pros/cons: *great nutritional value* <br> ... combines it all: an amazing product, quick and friendly service, cleanliness, great nutrition ... |
| pros/cons: *a bit pricey, healthy* <br> ... is an awesome place to go if you are health conscious. They have some really great low calorie dishes and they publish the calories and fat grams per serving. |

Figure 1: Excerpts from online restaurant reviews with pros/cons phrase lists. Both reviews assert that the restaurant serves healthy food, but use different keyphrases. Additionally, the first review discusses the restaurant's good service, but is not annotated as such in its keyphrases.

The major obstacle to the computational use of such free-text annotations is that they are inherently noisy — there is no fixed vocabulary, no explicit relationship between annotation keyphrases, and no guarantee that all relevant semantic properties of a document will be annotated. For example, in the pros/cons annotations accompanying the restaurant reviews in Figure 1, the same underlying semantic idea is expressed in different ways through the keyphrases "great nutritional value" and "healthy." Additionally, the first review discusses quality of service, but is not annotated as such. In contrast, expert annotations would replace synonymous keyphrases with a single canonical label, and would fully label all semantic properties described in the text. Such expert annotations are typically used in supervised learning methods. As we will demonstrate in the paper, traditional supervised approaches perform poorly when free-text annotations are used instead of clean, expert annotations.

This paper demonstrates a new approach for handling free-text annotation in the context of a hidden-topic analysis of the document text. We show that regularities in the text can clarify noise in the annotations — for example, although "great nutritional value" and "healthy" have different surface forms, the text in documents that are annotated by these two keyphrases will likely be similar. By modeling the relationship between document text and annotations over a large dataset, it is possible to induce a clustering over the annotation keyphrases that can help to overcome the problem of inconsistency. Our model also addresses the problem of incompleteness — when novice annotators fail to label relevant semantic topics — by estimating which topics are predicted by the document text alone.

Central to this approach is the idea that both document text and the associated annotations reflect a single underlying set of semantic properties. In the text, the semantic properties correspond to the induced hidden topics — this is similar to the growing body of work on latent topic models, such as latent Dirichlet allocation (LDA; Blei, Ng, & Jordan, 2003). However, unlike existing work on topic modeling, we tie hidden topics in the text with clusters of observed keyphrases. This connection is motivated by the idea that both the text and its associated annotations are grounded in a shared set of semantic properties. By modeling these properties directly, we ensure that the inferred hidden topics are semantically meaningful, and that the clustering over free-text annotations is robust to noise.

Our approach takes the form of a hierarchical Bayesian framework, and includes an LDA-style component in which each word in the text is generated from a mixture of multinomials. In addition, we also incorporate a similarity matrix across the universe of annotation keyphrases, which is constructed based on the orthographic and distributional features of the keyphrases. We model this matrix as being generated from an underlying clustering over the keyphrases, such that keyphrases that are clustered together are likely to produce high similarity scores. To generate the words in each document, we model two distributions over semantic properties — one governed by the annotation keyphrases and their clusters, and a background distribution to cover properties not mentioned in the annotations. The latent topic for each word is drawn from a mixture of these two distributions. After learning model parameters from a noisily-labeled training set, we can apply the model to unlabeled data.

We build a system that extracts semantic properties from reviews of products and services. This system uses as training corpus that includes user-created free-text annotations of the pros and cons in each review. Training yields two outputs: a clustering of keyphrases into semantic properties, and a topic model that is capable of inducing the semantic properties of unlabeled text. The clustering of annotation keyphrases is relevant for applications such as content-based information retrieval, allowing users to retrieve documents with semantically relevant annotations even if their surface forms differ from the query term. The topic model can be used to infer the semantic properties of unlabeled text.

The topic model can also be used to perform multi-document summarization, capturing the key semantic properties of multiple reviews. Unlike traditional extraction-based approaches to multi-document summarization, our induced topic model abstracts the text of each review into a representation capturing the relevant semantic properties. This enables comparison between reviews even when they use superficially different terminology to describe the same set of semantic properties. This idea is implemented in a review aggregation system that extracts the majority sentiment of multiple reviewers for each product or service. An example of the output produced by this system is shown in Figure 6. This system is applied to reviews in 480 product categories, allowing users to navigate the semantic properties of 49,490 products based on a total of 522,879 reviews. The effectiveness of our approach is confirmed by several evaluations.

For the summarization of both single and multiple documents, we compare the properties inferred by our model with expert annotations. Our approach yields substantially better results than alternatives from the research literature; in particular, we find that learning a clustering of free-text annotation keyphrases is essential to extracting meaningful semantic properties from our dataset. In addition, we compare the induced clustering with a gold standard clustering produced by expert annotators. The comparison shows that tying the clustering to the hidden topic model substantially improves its quality, and that the clustering induced by our system coheres well with the clustering produced by expert annotators.

The remainder of the paper is structured as follows. Section 2 compares our approach with previous work on topic modeling, semantic property extraction, and multi-document summarization. Section 3 describes the properties of free-text annotations that motivate our approach. The model itself is described in Section 4, and a method for parameter estimation is presented in Section 5. Section 6 describes the implementation and evaluation of single-document and multi-document summarization systems using these techniques. We summarize our contributions and consider directions for future work in Section 7. The code, datasets and expert annotations used in this paper are available online at `http://groups.csail.mit.edu/rbg/code/precis/`.

## 2. Related Work

The material presented in this section covers three lines of related work. First, we discuss work on Bayesian topic modeling that is related to our technique for learning from free-text annotations. Next, we discuss state-of-the-art methods for identifying and analyzing product properties from the review text. Finally, we situate our summarization work in the landscape of prior research on multi-document summarization.

### 2.1 Bayesian Topic Modeling

Recent work in the topic modeling literature has demonstrated that semantically salient topics can be inferred in an unsupervised fashion by constructing a generative Bayesian model of the document text. One notable example of this line of research is Latent Dirichlet Allocation (LDA; Blei et al., 2003). In the LDA framework, semantic topics are equated to latent distributions of words in a text; thus, each document is modeled as a mixture of topics. This class of models has been used for a variety of language processing tasks including topic segmentation (Purver, Körding, Griffiths, & Tenenbaum, 2006), named-entity resolution (Bhattacharya & Getoor, 2006), sentiment ranking (Titov & McDonald, 2008b), and word sense disambiguation (Boyd-Graber, Blei, & Zhu, 2007).

Our method is similar to LDA in that it assigns latent topic indicators to each word in the dataset, and models documents as mixtures of topics. However, the LDA model is unsupervised, and does not provide a method for linking the latent topics to external observed representations of the properties of interest. In contrast, our model exploits the free-text annotations in our dataset to ensure that the induced topics correspond to semantically meaningful properties.

Combining topics induced by LDA with external supervision was first considered by Blei and McAuliffe (2008) in their supervised Latent Dirichlet Allocation (sLDA) model. The induction of the hidden topics is driven by annotated examples provided during the training stage. From the perspective of supervised learning, this approach succeeds because the hidden topics mediate between document annotations and lexical features. Blei and McAuliffe describe a variational expectation-maximization procedure for approximate maximum-likelihood estimation of the model's parameters. When tested on two polarity assessment tasks, sLDA shows improvement over a model in which topics where induced by an unsupervised model and then added as features to a supervised model.

The key difference between our model and sLDA is that we do not assume access to clean supervision data during training. Since the annotations provided to our algorithm are free-text in nature, they are incomplete and fraught with inconsistency. This substantial difference in input structure motivates the need for a model that simultaneously induces the hidden structure in free-text annotations and learns to predict properties from text.

### 2.2 Property Assessment for Review Analysis

Our model is applied to the task of review analysis. Traditionally, the task of identifying the properties of a product from review texts has been cast as an extraction problem (Hu & Liu, 2004; Liu, Hu, & Cheng, 2005; Popescu, Nguyen, & Etzioni, 2005). For example, Hu and Liu (2004) employ association mining to identify noun phrases that express key portions of product reviews. The polarity of the extracted phrases is determined using a seed set of adjectives expanded via WordNet relations. A summary of a review is produced by extracting all property phrases present verbatim in the document.

Property extraction was further refined in OPINE (Popescu et al., 2005), another system for review analysis. OPINE employs a novel information extraction method to identify noun phrases that could potentially express the salient properties of reviewed products; these candidates are then pruned using WordNet and morphological cues. Opinion phrases are identified using a set of hand-crafted rules applied to syntactic dependencies extracted from the input document. The semantic orientation of properties is computed using a relaxation labeling method that finds the optimal assignment of polarity labels given a set of local constraints. Empirical results demonstrate that OPINE outperforms Hu and Liu's system in both opinion extraction and in identifying the polarity of opinion words.

These two feature extraction methods are informed by human knowledge about the way opinions are typically expressed in reviews: for Hu and Liu (2004), human knowledge is encoded using WordNet and the seed adjectives; for Popescu et al. (2005), opinion phrases are extracted via hand-crafted rules. An alternative approach is to learn the rules for feature extraction from annotated data. To this end, property identification can be modeled in a classification framework (Kim & Hovy, 2006). A classifier is trained using a corpus in which free-text pro and con keyphrases are specified by the review authors. These keyphrases are compared against sentences in the review text; sentences that exhibit high word overlap with previously identified phrases are marked as pros or cons according to the phrase polarity. The rest of the sentences are marked as negative examples.

Clearly, the accuracy of the resulting classifier depends on the quality of the automatically induced annotations. Our analysis of free-text annotations in several domains shows that automatically mapping from even manually-extracted annotation keyphrases to a document text is a difficult task, due to variability in keyphrase surface realizations (see Section 3). As we argue in the rest of this paper, it is beneficial to explicitly address the difficulties inherent in free-text annotations. To this end, our work is distinguished in two significant ways from the property extraction methods described above. First, we are able to predict properties beyond those that appear verbatim in the text. Second, our approach also learns the semantic relationships between different keyphrases, allowing us to draw direct comparisons between reviews even when the semantic ideas are expressed using different surface forms.

Working in the related domain of web opinion mining, Lu and Zhai (2008) describe a system that generates *integrated opinion summaries*, which incorporate expert-written articles (*e.g.,* a review from an online magazine) and user-generated "ordinary" opinion snippets (*e.g.,* mentions in blogs). Specifically, the expert article is assumed to be structured into segments, and a collection of representative ordinary opinions is aligned to each segment. Probabilistic Latent Semantic Analysis (PLSA) is used to induce a clustering of opinion snippets, where each cluster is attached to one of the expert article segments. Some clusters may also be unaligned to any segment, indicating opinions that are entirely unexpressed in the expert article. Ultimately, the integrated opinion summary is this combination of a single expert article with multiple user-generated opinion snippets that confirm or supplement specific segments of the review.

Our work's final goal is different — we aim to provide a highly compact summary of a multitude of user opinions by identifying the underlying semantic properties, rather than supplementing a single expert article with user opinions. We specifically leverage annotations that users already provide in their reviews, thus obviating the need for an expert article as a template for opinion inte-gration. Consequently, our approach is more suitable for the goal of producing concise keyphrase summarizations of user reviews, particularly when no review can be taken as authoritative.

The work closest in methodology to our approach is a review summarizer developed by Titov and McDonald (2008a). Their method summarizes a review by selecting a list of phrases that express writers' opinions in a set of predefined properties (*e.g.,*, *food* and *ambiance* for restaurant reviews). The system has access to numerical ratings in the same set of properties, but there is no training set providing examples of appropriate keyphrases to extract. Similar to sLDA, their method uses the numerical ratings to bias the hidden topics towards the desired semantic properties. Phrases that are strongly associated with properties via hidden topics are extracted as part of a summary.

There are several important differences between our work and the summarization method of Titov and McDonald. Their method assumes a predefined set of properties and thus cannot capture properties outside of that set. Moreover, consistent numerical annotations are required for training, while our method emphasizes the use of free-text annotations. Finally, since Titov and McDonald's algorithm is extractive, it does not facilitate property comparison across multiple reviews.

### 2.3 Multidocument Summarization

This paper also relates to a large body of work in multi-document summarization. Researchers have long noted that a central challenge of multi-document summarization is identifying redundant information over input documents (Radev & McKeown, 1998; Carbonell & Goldstein, 1998; Mani & Bloedorn, 1997; Barzilay, McKeown, & Elhadad, 1999). This task is of crucial significance because multi-document summarizers operate over related documents that describe the same facts multiple times. In fact, it is common to assume that repetition of information among related sources is an indicator of its importance (Barzilay et al., 1999; Radev, Jing, & Budzikowska, 2000; Nenkova, Vanderwende, & McKeown, 2006). Many of these algorithms first cluster sentences together, and then extract or generate sentence representatives for the clusters.

Identification of repeated information is equally central in our approach — our multi-document summarization method only selects properties that are stated by a plurality of users, thereby eliminating rare and/or erroneous opinions. The key difference between our algorithm and existing summarization systems is the method for identifying repeated expressions of a single semantic property. Since most of the existing work on multi-document summarization focuses on topic-independent newspaper articles, redundancy is identified via sentence comparison. For instance, Radev et al. (2000) compare sentences using cosine similarity between corresponding word vectors. Alternatively, some methods compare sentences via alignment of their syntactic trees (Barzilay et al., 1999; Marsi & Krahmer, 2005). Both string- and tree-based comparison algorithms are augmented with lexico-semantic knowledge using resources such as WordNet.

The approach described in this paper does not perform comparisons at the sentence level. Instead, we first abstract reviews into a set of properties and then compare property overlap across different documents. This approach relates to domain-dependent approaches for text summarization (Radev & McKeown, 1998; White, Korelsky, Cardie, Ng, Pierce, & Wagstaff, 2001; Elhadad & McKeown, 2001). These methods identify the relations between documents by comparing their abstract representations. In these cases, the abstract representation is constructed using off-the-shelf information extraction tools. A template specifying what types of information to select is crafted manually for a domain of interest. Moreover, the training of information extraction systems requires a corpus manually annotated with the relations of interest. In contrast, our method does not require manual template specification or corpora annotated by experts. While the abstract representations that we induce are not as linguistically rich as extraction templates, they nevertheless enable us to perform in-depth comparisons across different reviews.

| Property | Incompleteness | | | Inconsistency | |
|---|---|---|---|---|---|
| | Recall | Precision | F-score | Keyphrase Count | Top Keyphrase Coverage % |
| Good food | 0.736 | 0.968 | 0.836 | 23 | 38.3 |
| Good service | 0.329 | 0.821 | 0.469 | 27 | 28.9 |
| Good price | 0.500 | 0.707 | 0.586 | 20 | 41.8 |
| Bad food | 0.516 | 0.762 | 0.615 | 16 | 23.7 |
| Bad service | 0.475 | 0.633 | 0.543 | 20 | 22.0 |
| Bad price | 0.690 | 0.645 | 0.667 | 15 | 30.6 |
| Average | 0.578 | 0.849 | 0.688 | 22.6 | 33.6 |

Table 1: Incompleteness and inconsistency in the restaurant domain, for six major properties prevalent in the reviews. The incompleteness figures are the recall, precision, and F-score of the author annotations (manually clustered into properties) against the gold standard property annotations. Inconsistency is measured by the number of different keyphrase realizations with at least five occurrences associated with each property, and the percentage frequency with which the most commonly occurring keyphrases is used to annotate a property. The averages in the bottom row are weighted according to frequency of property occurrence.

## 3. Analysis of Free-Text Keyphrase Annotations

In this section, we explore the characteristics of free-text annotations, aiming to quantify the degree of noise observed in this data. The results of this analysis motivate the development of the learning algorithm described in Section 4.

We perform this investigation in the domain of online restaurant reviews using documents downloaded from the popular Epinions[1] website. Users of this website evaluate products by providing both a textual description of their opinion, as well as concise lists of keyphrases (pros and cons) summarizing the review. Pros/cons keyphrases are an appealing source of annotations for online review texts. However, they are contributed independently by multiple users and are thus unlikely to be as clean as expert annotations. In our analysis, we focus on two features of free-text annotations: *incompleteness* and *inconsistency*. The measure of incompleteness quantifies the degree of label omission in free-text annotations, while inconsistency reflects the variance of the keyphrase vocabulary used by various annotators.

To test the quality of these user-generated annotations, we compare them against "expert" annotations produced in a more systematic fashion. This annotation effort focused on six properties that were commonly mentioned by the review authors, specifically those shown in Table 1. Given a review and a property, the task is to assess whether the review's text supports the property. These annotations were produced by two judges guided by a standardized set of instructions. In contrast to author annotations from the website, the judges conferred during a training session to ensure consistency and completeness. The two judges collectively annotated 170 reviews, with 30 annotated

---

---

Property: *good price*
relatively inexpensive, dirt cheap, relatively cheap, great price, fairly priced, well priced, very reasonable prices, cheap prices, affordable prices, reasonable cost

---

Figure 2: Examples of the many different paraphrases related to the property *good price* that appear in the pros/cons keyphrases of reviews used for our inconsistency analysis.

by both. Cohen's Kappa, a measure of inter-annotator agreement that ranges from zero to one, is 0.78 on this joint set, indicating high agreement (Cohen, 1960). On average, each review text was annotated with 2.56 properties.

Separately, one of the judges also standardized the free-text pros/cons annotations for the same 170 reviews. Each review's keyphrases were matched to the same six properties. This standardization allows for direct comparison between the properties judged to be supported by a review's text and the properties described in the same review's free-text annotations. We find that many semantic properties that were judged to be present in the text were not user annotated — on average, the keyphrases expressed 1.66 relevant semantic properties per document, while the text expressed 2.56 properties. This gap demonstrates the frequency with which authors omitted relevant semantic properties from their review annotations.

### 3.1 Incompleteness

To measure incompleteness, we compare the properties stated by review authors in the form of pros and cons against those stated only in the review text, as judged by expert annotators. This comparison is performed using precision, recall and F-score. In this setting, recall is the proportion of semantic properties in the text for which the review author also provided at least one annotation keyphrase; precision is the proportion of keyphrases that conveyed properties judged to be supported by the text; and F-score is their harmonic mean. The results of the comparison are summarized in the left half of Table 1.

These incompleteness results demonstrate the significant discrepancy between user and expert annotations. As expected, recall is quite low; more than 40% of property occurrences are stated in the review text without being explicitly mentioned in the annotations. The precision scores indicate that the converse is also true, though to a lesser extent — some keyphrases will express properties not mentioned in text.

Interestingly, precision and recall vary greatly depending on the specific property. They are highest for *good food*, matching the intuitive notion that high food quality would be a key salient property of a restaurant, and thus more likely to be mentioned in both text and annotations. Conversely, the recall for *good service* is lower — for most users, high quality of service is apparently not a key point when summarizing a review with keyphrases.

### 3.2 Inconsistency

The lack of a unified annotation scheme in the restaurant review dataset is apparent — across all reviewers, the annotations feature 26,801 unique keyphrase surface forms over a set of 49,310 total keyphrase occurrences. Clearly, many unique keyphrases express the same semantic property — in Figure 2, *good price* is expressed in ten different ways. To quantify this phenomenon, the judges manually clustered a subset of the keyphrases associated with the six previously mentioned properties. Specifically, 121 keyphrases associated with the six major properties were chosen, accounting for 10.8% of all keyphrase occurrences.

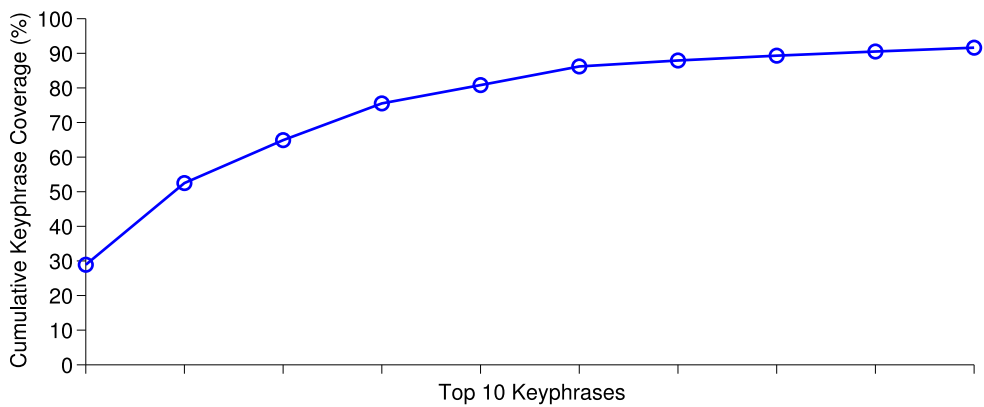

Figure 3: Cumulative occurrence counts for the top ten keyphrases associated with the *good service* property. The percentages are out of a total of 1,210 separate keyphrase occurrences for this property.

We use these manually clustered annotations to examine the distributional pattern of keyphrases that describe the same underlying property, using two different statistics. First, the number of different keyphrases for each property gives a lower bound on the number of possible paraphrases. Second, we measure how often the most common keyphrase is used to annotate each property, *i.e.,* the *coverage* of that keyphrase. This metric gives a sense of how diffuse the keyphrases within a property are, and specifically whether one single keyphrase dominates occurrences of the property. Note that this value is an overestimate of the true coverage, since we are only considering a tenth of all keyphrase occurrences.

The right half of Table 1 summarizes the variability of property paraphrases. Observe that each property is associated with numerous paraphrases, all of which were found multiple times in the actual keyphrase set. Most importantly, the most frequent keyphrase accounted for only about a third of all property occurrences, strongly suggesting that targeting only these labels for learning is a very limited approach. To further illustrate this last point, consider the property of *good service*, whose keyphrase realizations' distributional histogram appears in Figure 3. The cumulative percentage frequencies of the most frequent keyphrases associated with this property are plotted. The top four keyphrases here account for only three quarters of all property occurrences, even within the limited set of keyphrases we consider in this analysis, motivating the need for aggregate consideration of keyphrases.

In the next section, we introduce a model that induces a clustering among keyphrases while relating keyphrase clusters to the text, directly addressing these characteristics of the data.

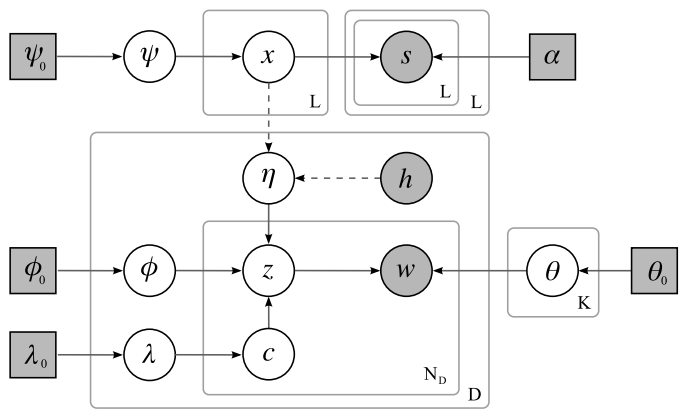

$\psi$  –  keyphrase cluster model

$x$  –  keyphrase cluster assignment

$s$  –  keyphrase similarity values

$h$  –  document keyphrases

$\eta$  –  document keyphrase topics

$\lambda$  –  probability of selecting $\eta$ instead of $\phi$

$c$  –  selects between $\eta$ and $\phi$ for word topics

$\phi$  –  background word topic model

$z$  –  word topic assignment

$\theta$  –  language models of each topic

$w$  –  document words

$$\psi \sim \text{Dirichlet}(\psi_0)$$

$$x_\ell \sim \text{Multinomial}(\psi)$$

$$s_{\ell,\ell'} \sim \begin{cases} \text{Beta}(\alpha_=) & \text{if } x_\ell = x_{\ell'} \\ \text{Beta}(\alpha_{\neq}) & \text{otherwise} \end{cases}$$

$$\eta_d = [\eta_{d,1} \dots \eta_{d,K}]^\text{T}$$

$$\text{where } \eta_{d,k} \propto \begin{cases} 1 & \text{if } x_\ell = k \text{ for any } l \in h_d \\ \epsilon & \text{otherwise} \end{cases}$$

$$\lambda_d \sim \text{Beta}(\lambda_0)$$

$$c_{d,n} \sim \text{Bernoulli}(\lambda_d)$$

$$\phi_d \sim \text{Dirichlet}(\phi_0)$$

$$z_{d,n} \sim \begin{cases} \text{Multinomial}(\eta_d) & \text{if } c_{d,n} = 1 \\ \text{Multinomial}(\phi_d) & \text{otherwise} \end{cases}$$

$$\theta_k \sim \text{Dirichlet}(\theta_0)$$

$$w_{d,n} \sim \text{Multinomial}(\theta_{z_{d,n}})$$

Figure 4: The plate diagram for our model. Shaded circles denote observed variables, and squares denote hyperparameters. The dotted arrows indicate that $\eta$ is constructed deterministically from $x$ and $h$. We use $\epsilon$ to refer to a small constant probability mass.

## 4. Model Description

We present a generative Bayesian model for documents annotated with free-text keyphrases. Our model assumes that each annotated document is generated from a set of underlying semantic *topics*. Semantic topics generate the document text by indexing a language model; in our approach, they are also associated with clusters of keyphrases. In this way, the model can be viewed as an extension of Latent Dirichlet Allocation (Blei et al., 2003), where the latent topics are additionally biased toward the keyphrases that appear in the training data. However, this coupling is flexible, as some words are permitted to be drawn from topics that are not represented by the keyphrase annotations. This permits the model to learn effectively in the presence of incomplete annotations, while still encouraging the keyphrase clustering to cohere with the topics supported by the document text.

Another critical aspect of our model is that we desire the ability to use arbitrary comparisons between keyphrases, in addition to information about their surface forms. To accommodate this goal, we do not treat the keyphrase surface forms as generated from the model. Rather, we acquire a real-valued similarity matrix across the universe of possible keyphrases, and treat this matrix as generated from the keyphrase clustering. This representation permits the use of surface and distributional features for keyphrase similarity, as described in Section 4.1.

An advantage of hierarchical Bayesian models is that it is easy to change which parts of the model are observed and which parts are hidden. During training, the keyphrase annotations are observed, so that the hidden semantic topics are coupled with clusters of keyphrases. To account for words not related to semantic topics, some topics may not have any associated keyphrases. At test time, the model is presented with documents for which the keyphrase annotations are hidden. The model is evaluated on its ability to determine which keyphrases are applicable, based on the hidden topics present in the document text.

The judgment of whether a topic applies to a given unannotated document is based on the probability mass assigned to that topic in the document's background topic distribution. Because there are no annotations, the background topic distribution should capture the entirety of the document's topics. For the task involving reviews of products and services, multiple topics may accompany each document. In this case, each topic whose probability is above a threshold (tuned on the development set) is predicted as being supported.

### 4.1 Keyphrase Clustering

To handle the hidden paraphrase structure of the keyphrases, one component of the model estimates a clustering over keyphrases. The goal is to obtain clusters where each cluster correspond to a well-defined semantic topic — *e.g.,* both "healthy" and "good nutrition" should be grouped into a single cluster. Because our overall joint model is generative, a generative model for clustering could easily be integrated into the larger framework. Such an approach would treat all of the keyphrases in each cluster as being generated from a parametric distribution. However, this representation would not permit many powerful features for assessing the similarity of pairs of keyphrases, such as string overlap or keyphrase co-occurrence in a corpus (McCallum, Bellare, & Pereira, 2005).

For this reason, we represent each keyphrase as a real-valued vector rather than as its surface form. The vector for a given keyphrase includes the similarity scores with respect to every other observed keyphrase (the similarity scores are represented by $s$ in Figure 4). We model these similarity scores as generated by the cluster memberships (represented by $x$ in Figure 4). If two keyphrases are clustered together, their similarity score is generated from a distribution encouraging high similarity; otherwise, a distribution encouraging low similarity is used.[2]

| Lexical | The cosine similarity between the surface forms of two keyphrases, represented as word frequency vectors. |
|---|---|
| Co-occurrence | Each keyphrase is represented as a vector of co-occurrence values. This vector counts how many times other keyphrases appear in documents annotated with this keyphrase. For example, the similarity vector for "good food" may include an entry for "very tasty food," the value of which would be the number of documents annotated with "good food" that contain "very tasty food" in their text. The similarity between two keyphrases is then the cosine similarity of their co-occurrence vectors. |

Table 2: The two sources of information used to compute the similarity matrix for our experiments. The final similarity scores are linear combinations of these two values. Note that co-occurrence similarity contains second-order co-occurrence information.

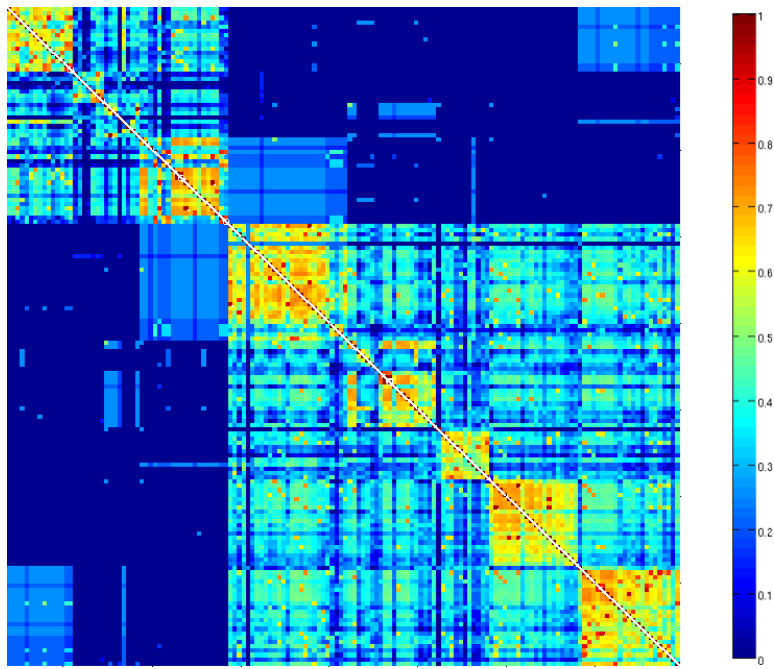

Figure 5: A surface plot of the keyphrase similarity matrix from a set of restaurant reviews, computed according to Table 2. Red indicates high similarity, whereas blue indicates low similarity. In this diagram, the keyphrases have been grouped according to an expert-created clustering, so keyphrases of similar meaning are close together. The strong series of similarity "blocks" along the diagonal hint at how this information could induce a reasonable clustering.

The features used for producing the similarity matrix are given in Table 2, encompassing lexical and distributional similarity measures. Our implemented system takes a linear combination of these two data sources, weighting both sources equally. The resulting similarity matrix for keyphrases from the restaurant domain is shown in Figure 5.

As described in the next section, when clustering keyphrases, our model takes advantage of the topic structure of documents annotated with those keyphrases, in addition to information about the individual keyphrases themselves. In this sense, it differs from traditional approaches for paraphrase identification (Barzilay & McKeown, 2001; Lin & Pantel, 2001).

## 4.2 Document Topic Modeling

Our analysis of the document text is based on probabilistic topic models such as LDA (Blei et al., 2003). In the LDA framework, each word is generated from a language model that is indexed by the word's topic assignment. Thus, rather than identifying a single topic for a document, LDA identifies a distribution over topics. High probability topic assignments will identify compact, low-entropy language models, so that the probability mass of the language model for each topic is divided among a relatively small vocabulary.

Our model operates in a similar manner, identifying a topic for each word, denoted by $z$ in Figure 4. However, where LDA learns a distribution over topics for each document, we deterministically construct a document-specific topic distribution from the clusters represented by the document's keyphrases — this is $\eta$ in the figure. $\eta$ assigns equal probability to all topics that are represented in the keyphrase annotations, and very small probability to other topics. Generating the word topics in this way ties together the clustering and language models.

As noted above, sometimes the keyphrase annotation does not represent all of the semantic topics that are expressed in the text. For this reason, we also construct another "background" distribution $\phi$ over topics. The auxiliary variable $c$ indicates whether a given word's topic is drawn from the distribution derived from annotations, or from the background model. Representing $c$ as a hidden variable allows us to stochastically interpolate between the two language models $\phi$ and $\eta$. In addition, any given document will most likely also discuss topics that are not covered by any keyphrase. To account for this, the model is allowed to leave some of the clusters empty, thus leaving some of the topics to be independent of all the keyphrases.

## 4.3 Generative Process

Our model assumes that all observed data is generated through a stochastic process involving hidden parameters. In this section, we formally specify this generative process. This specification guides inference of the hidden parameters based on observed data, which are the following:

- For each of the $L$ keyphrases, a vector $\mathbf{s}_\ell$ of length $L$ denoting a pairwise similarity score in the interval $[0, 1]$ to every other keyphrase.

- For each document $d$, its bag of words $\mathbf{w}_d$ of length $N_d$. The $n$th word of $d$ is $w_{d,n}$.

---

2. Note that while we model each similarity score as an independent draw; clearly this assumption is too strong, due to symmetry and transitivity. Models making similar assumptions about the independence of related hidden variables have previously been shown to be successful (for example, Toutanova & Johnson, 2008).

- For each document $d$, a set of keyphrase annotations $h_d$, which includes index $\ell$ if the document was annotated with keyphrase $\ell$.

- The number of clusters $K$, which should be large enough to encompass topics with actual clusters of keyphrases, as well as word-only topics.

These observed variables are generated according to the following process:

1. Draw a multinomial distribution $\psi$ over the $K$ keyphrase clusters from a symmetric Dirichlet prior with parameter $\psi_0$.[3]

2. For $\ell = 1 \ldots L$:

   (a) Draw the $\ell$th keyphrase's cluster assignment $x_\ell$ from Multinomial($\psi$).

3. For $(\ell, \ell') = (1 \ldots L, 1 \ldots L)$:

   (a) If $x_\ell = x_{\ell'}$, draw $s_{\ell,\ell'}$ from Beta($\alpha_=$) $\equiv$ Beta(2, 1), encouraging scores to be biased toward values close to one.

   (b) If $x_\ell \neq x_{\ell'}$, draw $s_{\ell,\ell'}$ from Beta($\alpha_{\neq}$) $\equiv$ Beta(1, 2), encouraging scores to be biased toward values close to zero.

4. For $k = 1 \ldots K$:

   (a) Draw language model $\theta_k$ from a symmetric Dirichlet prior with parameter $\theta_0$.

5. For $d = 1 \ldots D$:

   (a) Draw a background topic model $\phi_d$ from a symmetric Dirichlet prior with parameter $\phi_0$.

   (b) Deterministically construct an annotation topic model $\eta_d$, based on keyphrase cluster assignments $\mathbf{x}$ and observed document annotations $h_d$. Specifically, let $\mathbf{H}$ be the set of topics represented by phrases in $h_d$. Distribution $\eta_d$ assigns equal probability to each element of $\mathbf{H}$, and a very small probability mass to other topics.[4]

   (c) Draw a weighted coin $\lambda_d$ from Beta($\lambda_0$), which will determine the balance between annotation $\eta_d$ and background topic models $\phi_d$.

   (d) For $n = 1 \ldots N_d$:

      i. Draw a binary auxiliary variable $c_{d,n}$ from Bernoulli($\lambda_d$), which determines whether the topic of the word $w_{d,n}$ is drawn from the annotation topic model $\eta_d$ or the background model $\phi_d$.

      ii. Draw a topic assignment $z_{d,n}$ from the appropriate multinomial as indicated by $c_{d,n}$.

      iii. Draw word $w_{d,n}$ from Multinomial($\theta_{z_{d,n}}$), that is, the language model indexed by the word's topic.

---

3. Variables subscripted with zero are fixed hyperparameters.

4. Making a hard assignment of zero probability to the other topics creates problems for parameter estimation. A probability of $10^{-4}$ was assigned to all topics not represented by the keyphrase cluster memberships.

## 5. Parameter Estimation

To make predictions on unseen data, we need to estimate the parameters of the model. In Bayesian inference, we estimate the distribution for each parameter, conditioned on the observed data and hyperparameters. Such inference is intractable in the general case, but sampling approaches allow us to approximately construct distributions for each parameter of interest.

Gibbs sampling is perhaps the most generic and straightforward sampling technique. Conditional distributions are computed for each hidden variable, given all the other variables in the model. By repeatedly sampling from these distributions in turn, it is possible to construct a Markov chain whose stationary distribution is the posterior of the model parameters (Gelman, Carlin, Stern, & Rubin, 2004). The use of sampling techniques in natural language processing has been previously investigated by many researchers, including Finkel, Grenager, and Manning (2005) and Goldwater, Griffiths, and Johnson (2006).

We now present sampling equations for each of the hidden variables in Figure 4. The prior over keyphrase clusters $\psi$ is sampled based on the hyperprior $\psi_0$ and the keyphrase cluster assignments $\mathbf{x}$. We write $p(\psi \mid \ldots)$ to mean the probability conditioned on all the other variables.

$$
\begin{aligned}
p(\psi \mid \ldots) &\propto p(\psi \mid \psi_0)p(\mathbf{x} \mid \psi), \\
&= p(\psi \mid \psi_0) \prod_\ell p(x_\ell \mid \psi) \\
&= \text{Dirichlet}(\psi; \psi_0) \prod_\ell \text{Multinomial}(x_\ell; \psi) \\
&= \text{Dirichlet}(\psi; \psi'),
\end{aligned}
$$

where $\psi'_i$ is $\psi_0 + \text{count}(x_\ell = i)$. This conditional distribution is derived based on the conjugacy of the multinomial to the Dirichlet distribution. The first line follows from Bayes' rule, and the second line from the conditional independence of cluster assignments $\mathbf{x}$ given keyphrase distribution $\psi$.

Resampling equations for $\phi_d$ and $\theta_k$ can be derived in a similar manner:

$$
\begin{aligned}
p(\phi_d \mid \ldots) &\propto \text{Dirichlet}(\phi_d; \phi'_d), \\
p(\theta_k \mid \ldots) &\propto \text{Dirichlet}(\theta_k; \theta'_k),
\end{aligned}
$$

where $\phi'_{d,i} = \phi_0 + \text{count}(z_{n,d} = i \wedge c_{n,d} = 0)$ and $\theta'_{k,i} = \theta_0 + \sum_d \text{count}(w_{n,d} = i \wedge z_{n,d} = k)$. In building the counts for $\phi'_i$, we consider only cases in which $c_{n,d} = 0$, indicating that the topic $z_{n,d}$ is indeed drawn from the background topic model $\phi_d$. Similarly, when building the counts for $\theta'_k$, we consider only cases in which the word $w_{d,n}$ is drawn from topic $k$.

To resample $\lambda$, we employ the conjugacy of the Beta prior to the Bernoulli observation likelihoods, adding counts of $\mathbf{c}$ to the prior $\lambda_0$.

$$
p(\lambda_d \mid \ldots) \propto \text{Beta}(\lambda_d; \lambda'_d),
$$

where $\lambda'_d = \lambda_0 + \left[ \begin{array}{c} \sum_n \text{count}(c_{d,n} = 1) \\ \sum_n \text{count}(c_{d,n} = 0) \end{array} \right]$.

The keyphrase cluster assignments are represented by $\mathbf{x}$, whose sampling distribution depends on $\psi$, $\mathbf{s}$, and $\mathbf{z}$, via $\eta$:

$$p(x_\ell \mid \dots) \propto p(x_\ell \mid \psi)p(\mathbf{s} \mid x_\ell, \mathbf{x}_{-\ell}, \alpha)p(\mathbf{z} \mid \eta, \psi, \mathbf{c})$$

$$\propto p(x_\ell \mid \psi)\left[\prod_{\ell' \neq \ell} p(s_{\ell,\ell'} \mid x_\ell, x_{\ell'}, \alpha)\right]\left[\prod_d^D \prod_{c_{d,n}=1} p(z_{d,n} \mid \eta_d)\right]$$

$$= \text{Multinomial}(x_\ell; \psi)\left[\prod_{\ell' \neq \ell} \text{Beta}(s_{\ell,\ell'}; \alpha_{x_\ell,x_{\ell'}})\right]\left[\prod_d^D \prod_{c_{d,n}=1} \text{Multinomial}(z_{d,n}; \eta_d)\right].$$

The leftmost term of the above equation is the prior on $x_\ell$. The next term encodes the dependence of the similarity matrix $\mathbf{s}$ on the cluster assignments; with slight abuse of notation, we write $\alpha_{x_\ell,x_{\ell'}}$ to denote $\alpha_=$ if $x_\ell = x_{\ell'}$, and $\alpha_{\neq}$ otherwise. The third term is the dependence of the word topics $z_{d,n}$ on the topic distribution $\eta_d$. We compute the final result of this probability expression for each possible setting of $x_\ell$, and then sample from the normalized multinomial.

The word topics $\mathbf{z}$ are sampled according to the topic distribution $\eta_d$, the background distribution $\phi_d$, the observed words $\mathbf{w}$, and the auxiliary variable $\mathbf{c}$:

$$p(z_{d,n} \mid \dots) \propto p(z_{d,n} \mid \phi, \eta_d, c_{d,n})p(w_{d,n} \mid z_{d,n}, \theta)$$

$$= \begin{cases} \text{Multinomial}(z_{d,n}; \eta_d)\text{Multinomial}(w_{d,n}; \theta_{z_{d,n}}) & \text{if } c_{d,n} = 1, \\ \text{Multinomial}(z_{d,n}; \phi_d)\text{Multinomial}(w_{d,n}; \theta_{z_{d,n}}) & \text{otherwise.} \end{cases}$$

As with $x$, each $z_{d,n}$ is sampled by computing the conditional likelihood of each possible setting within a constant of proportionality, and then sampling from the normalized multinomial.

Finally, we sample the auxiliary variable $c_{d,n}$, which indicates whether the hidden topic $z_{d,n}$ is drawn from $\eta_d$ or $\phi_d$. $\mathbf{c}$ depends on its prior $\lambda$ and the hidden topic assignments $\mathbf{z}$:

$$p(c_{d,n} \mid \dots) \propto p(c_{d,n} \mid \lambda_d)p(z_{d,n} \mid \eta_d, \phi_d, c_{d,n})$$

$$= \begin{cases} \text{Bernoulli}(c_{d,n}; \lambda_d)\text{Multinomial}(z_{d,n}; \eta_d) & \text{if } c_{d,n} = 1, \\ \text{Bernoulli}(c_{d,n}; \lambda_d)\text{Multinomial}(z_{d,n}; \phi_d) & \text{otherwise.} \end{cases}$$

Again, we compute the likelihood of $c_{d,n} = 0$ and $c_{d,n} = 1$ within a constant of proportionality, and then sample from the normalized Bernoulli distribution.

Finally, our model requires values for fixed hyperparameters $\theta_0$, $\lambda_0$, $\psi_0$, and $\phi_0$, which are tuned in the standard way based on development set performance. Appendix C lists the hyperparameters values used for each domain in our experiments.

One of the main applications of our model is to predict the properties supported by documents that are not annotated with keyphrases. At test time, we would like to compute a posterior estimate of $\phi_d$ for an unannotated test document $d$. Since annotations are not present, property prediction is based only on the text component of the model. For this estimate, we use the same Gibbs sampling procedure, restricted to $z_{d,n}$ and $\phi_d$, with the stipulation that $c_{d,n}$ is fixed at zero so that $z_{d,n}$ is always drawn from $\phi_d$. In particular, we treat the language models as known; to more accurately integrate over all possible language models, we use the final 1000 samples of the language models from training as opposed to using a point estimate. For each topic, if its probability in $\phi_d$ exceeds a certain threshold, that topic is predicted. This threshold is tuned independently for each topic on a development set. The empirical results we present in Section 6 are obtained in this manner.

Figure 6: Summary of reviews for the movie *Pirates of the Caribbean: At World's End* on Précis. This summary is based on 27 documents. The list of pros and cons are generated automatically using the system described in this paper. The generation of numerical ratings is based on the algorithm described in Snyder and Barzilay (2007).

# 6. Evaluation of Summarization Quality

Our model for document analysis is implemented in Précis,[5] a system that performs single- and multi-document review summarization. The goal of Précis is to provide users with effective access to review data via mobile devices. Précis contains information about 49,490 products and services ranging from childcare products to restaurants and movies. For each of these products, the system contains a collection of reviews downloaded from consumer websites such as Epinions, CNET, and Amazon. Précis compresses data for each product into a short list of pros and cons that are supported by the majority of reviews. An example of a summary of 27 reviews for the movie *Pirates of the Caribbean: At World's End* is shown in Figure 6. In contrast to traditional multi-document summarizers, the output of the system is not a sequence of sentences, but rather a list of phrases indicative of product properties. This summarization format follows the format of pros/cons summaries that individual reviewers provide on multiple consumer websites. Moreover, the brevity of the summary is particularly suitable for presenting on small screens such as those of mobile devices.

To automatically generate the combined pros/cons list for a product or service, we first apply our model to each review. The model is trained independently for each product domain (*e.g.,* movies) using a corresponding subset of reviews with free-text annotations. These annotations also provide a set of keyphrases that contribute to the clusters associated with product properties. Once the model is trained, it labels each review with a set of properties. Since the set of possible properties is the same for all reviews of a product, the comparison among reviews is straightforward — for each property, we count the number of reviews that support it, and select the property as part of a summary if it is supported by the majority of the reviews. The set of semantic properties is converted into a pros/cons list by presenting the most common keyphrase for each property.

---

5. Précis is accessible at `http://groups.csail.mit.edu/rbg/projects/precis/`.

This aggregation technology is applicable in two scenarios. The system can be applied to unannotated reviews, inducing semantic properties from the document text; this conforms to the traditional way in which learning-based systems are applied to unlabeled data. However, our model is valuable even when individual reviews do include pros/cons keyphrase annotations. Due to the high degree of paraphrasing, direct comparison of keyphrases is challenging (see Section 3). By inferring a clustering over keyphrases, our model permits comparison of keyphrase annotations on a more semantic level.

The remainder of this section provides a set of evaluations of our model's ability to capture the semantic content of document text and keyphrase annotations. Section 6.1 describes an evaluation of our system's ability to extract meaningful semantic summaries from individual documents, and also assesses the quality of the paraphrase structure induced by our model. Section 6.2 extends this evaluation to our system's ability to summarize multiple review documents.

## 6.1 Single-Document Evaluation

First, we evaluate our model with respect to its ability to reproduce the annotations present in individual documents, based on the document text. We compare against a wide variety of baselines and variations of our model, demonstrating the appropriateness of our approach to this task. In addition, we explicitly evaluate the quality of the paraphrase structure induced by our model by comparing against a gold standard clustering of keyphrases provided by expert annotators.

### 6.1.1 EXPERIMENTAL SETUP

In this section, we describe the datasets and evaluation techniques used for experiments with our system and other automatic methods. We also comment on how hyperparameters are tuned for our model, and how sampling is initialized.

| Statistic | Restaurants | Cell Phones | Digital Cameras |
|---|---|---|---|
| # of reviews | 5735 | 1112 | 3971 |
| avg. review length | 786.3 | 1056.9 | 1014.2 |
| avg. keyphrases / review | 3.42 | 4.91 | 4.84 |

Table 3: Statistics of the datasets used in our evaluations

**Data Sets** We evaluate our system on reviews from three domains: restaurants, cell phones, and digital cameras. These reviews were downloaded from the Epinions website; we used user-authored pros and cons associated with reviews as keyphrases (see Section 3). Statistics for the datasets are provided in Table 3. For each of the domains, we selected 50% of the documents for training.

We consider two strategies for constructing test data. First, we consider evaluating the semantic properties inferred by our system against expert annotations of the semantic properties present in each document. To this end, we use the expert annotations originally described in Section 3 as a test set;[6] to reiterate, these were annotations of 170 reviews in the restaurant domain, of which we now hold out 50 as a development set. The review texts were annotated with six properties according to standardized guidelines. This strategy enforces consistency and completeness in the ground truth annotations, differentiating them from free-text annotations.

Unfortunately, our ability to evaluate against expert annotations is limited by the cost of producing such annotations. To expand evaluation to other domains, we use the author-written keyphrase annotations that are present in the original reviews. Such annotations are noisy — while the presence of a property annotation on a document is strong evidence that the document supports the property, the inverse is not necessarily true. That is, the *lack* of an annotation does not necessarily imply that its respective property does not hold — *e.g.,* a review with no *good service*-related keyphrase may still praise the service in the body of the document.

For experiments using free-text annotations, we overcome this pitfall by restricting the evaluation of predictions of individual properties to only those documents that are annotated with that property or its antonym. For instance, when evaluating the prediction of the *good service* property, we will only select documents which are either annotated with *good service* or *bad service*-related keyphrases.[7] For this reason, each semantic property is evaluated against a unique subset of documents. The details of these development and test sets are presented in Appendix A.

To ensure that free-text annotations can be reliably used for evaluation, we compare with the results produced on expert annotations whenever possible. As shown in Section 6.1.2, the free-text evaluations produce results that cohere well with those obtained on expert annotations, suggesting that such labels can be used as a reasonable proxy for expert annotation evaluations.

**Evaluation Methods**    Our first evaluation leverages the expert annotations described in Section 3. One complication is that expert annotations are marked on the level of semantic properties, while the model makes predictions about the appropriateness of individual keyphrases. We address this by representing each expert annotation with the most commonly-observed keyphrase from the manually-annotated cluster of keyphrases associated with the semantic property. For example, an annotation of the semantic property *good food* is represented with its most common keyphrase realization, "great food." Our evaluation then checks whether this keyphrase is within any of the clusters of keyphrases predicted by the model.

The evaluation against author free-text annotations is similar to the evaluation against expert annotations. In this case, the annotation takes the form of individual keyphrases rather than semantic properties. As noted, author-generated keyphrases suffer from inconsistency. We obtain a consistent evaluation by mapping the author-generated keyphrase to a cluster of keyphrases as a determined by the expert annotator, and then again selecting the most common keyphrase realization of the cluster. For example, the author may use the keyphrase "tasty," which maps to the semantic cluster *good food*; we then select the most common keyphrase realization, "great food." As in the expert evaluation, we check whether this keyphrase is within any of the clusters predicted by the model.

Model performance is quantified using recall, precision, and F-score. These are computed in the standard manner, based on the model's representative keyphrase predictions compared against the corresponding references. Approximate randomization (Yeh, 2000; Noreen, 1989) is used for statistical significance testing. This test repeatedly performs random swaps of individual results from each candidate system, and checks whether the resulting performance gap remains at least as large. We use this test because it is valid for comparing nonlinear functions of random variables, such as F-scores, unlike other common methods such as the sign test. Previous work that used this test include evaluations at the Message Understanding Conference (Chinchor, Lewis, & Hirschman, 1993; Chinchor, 1995); more recently, Riezler and Maxwell (2005) advocated for its use in evaluating machine translation systems.

---

6. The expert annotations are available at http://groups.csail.mit.edu/rbg/code/precis/.

7. This determination is made by mapping author keyphrases to properties using an expert-generated gold standard clustering of keyphrases. It is much cheaper to produce an expert clustering of keyphrases than to obtain expert annotations of the semantic properties in every document.

**Parameter Tuning and Initialization**   To improve the model's convergence rate, we perform two initialization steps for the Gibbs sampler. First, sampling is done only on the keyphrase clustering component of the model, ignoring document text. Second, we fix this clustering and sample the remaining model parameters. These two steps are run for 5,000 iterations each. The full joint model is then sampled for 100,000 iterations. Inspection of the parameter estimates confirms model convergence. On a 2GHz dual-core desktop machine, a multithreaded C++ implementation of model training takes about two hours for each dataset.

Our model needs to be provided with the number of clusters $K$.[8] We set $K$ large enough for the model to learn effectively on the development set. For the restaurant data we set $K$ to 20. For cell phones and digital cameras, $K$ was set to 30 and 40, respectively. These values were tuned using the development set. However, we found that as long as $K$ was large enough to accommodate a significant number of keyphrase clusters, and a few additional to account for topics with no keyphrases, the specific value of $K$ does not affect the model's performance. All other hyperparameters were adjusted based on development set performance, though tuning was not extensive.

As previously mentioned, we obtain document properties by examining the probability mass of the topic distribution assigned to each property. A probability threshold is set for each property via the development set, optimizing for maximum F-score.

### 6.1.2 RESULTS

In this section, we report the performance of our model, comparing it with an array of increasingly sophisticated baselines and model variations. We first demonstrate that learning a clustering of annotation keyphrases is crucial for accurate semantic prediction. Next, we investigate the impact of paraphrasing quality on model accuracy by considering the expert-generated gold standard clustering of keyphrases as another comparison point; we also consider alternative automatically computed sources of paraphrase information.

For ease of comparison, the results of all the experiments are shown in Table 5 and Table 6, with a summary of the baselines and model variations in Table 4.

**Comparison against Simple Baselines**   Our first evaluation compares our model to four naïve baselines. All four treat keyphrases as independent, ignoring their latent paraphrase structure.

- *Random:* Each keyphrase is supported by a document with probability of one half. The results of this baseline are computed in expectation, rather than actually run. This baseline is expected to have a recall of 0.5, because in expectation it will select half of the correct keyphrases. Its precision is the average proportion of annotations in the test set against the number of possible annotations. That is, in a test set of size $n$ with $m$ properties, if property

---

8. This requirement could conceivably be removed by modeling the cluster indices as being drawn from a Dirichlet process prior.

| | |
|---|---|
| Random | Each keyphrase is supported by a document with probability of one half. |
| Keyphrase in text | A keyphrase is supported by a document if it appears verbatim in the text. |
| Keyphrase classifier | A separate support vector machine classifier is trained for each keyphrase. Positive examples are documents that are labeled by the author with the keyphrase; all other documents are considered to be negative examples. A keyphrase is supported by a document if that keyphrase's classifier returns a positive prediction. |
| Heuristic keyphrase classifier | Similar to *keyphrase classifier*, except heuristic methods are used in an attempt to reduce noise from the training documents. Specifically we wish to remove sentences that discuss other keyphrases from the positive examples. The heuristic removes from the positive examples all sentences that have no word overlap with the given keyphrase. |
| Model cluster in text | A keyphrase is supported by a document if it or any of its paraphrases appear in the text. Paraphrasing is based on our model's keyphrase clusters. |
| Model cluster classifier | A separate classifier is trained for each cluster of keyphrases. Positive examples are documents that are labeled by the author with any keyphrase from the cluster; all other documents are negative examples. All keyphrases of a cluster are supported by a document if that cluster's classifier returns a positive prediction. Keyphrase clustering is based on our model. |
| Heuristic model cluster classifier | Similar to *model cluster classifier*, except heuristic methods are used to reduce noise from the training documents. Specifically we wish to remove from the positive examples sentences that discuss keyphrases from other clusters. The heuristic removes from the positive examples all sentences that have no word overlap with any of the keyphrases from the given cluster. Keyphrase clustering is based on our model. |
| Gold cluster model | A variation of our model where the clustering of keyphrases is fixed to an expert-created gold standard. Only the text modeling parameters are learned. |
| Gold cluster in text | Similar to *model cluster in text*, except the clustering of keyphrases is according to the expert-produced gold standard. |
| Gold cluster classifier | Similar to *model cluster classifier*, except the clustering of keyphrases is according to the expert-produced gold standard. |
| Heuristic gold cluster classifier | Similar to *heuristic model cluster classifier*, except the clustering of keyphrases is according to the expert-produced gold standard. |
| Independent cluster model | A variation of our model where the clustering of keyphrases is first learned from keyphrase similarity information only, separately from the text. The resulting *independent* clustering is then fixed while the text modeling parameters are learned. This variation's key distinction from our full model is the lack of joint learning of keyphrase clustering and text topics. |
| Independent cluster in text | Similar to *model cluster in text*, except that the clustering of keyphrases is according to the independent clustering. |
| Independent cluster classifier | Similar to *model cluster classifier*, except that the clustering of keyphrases is according to the independent clustering. |
| Heuristic independent cluster classifier | Similar to *heuristic model cluster classifier*, except the clustering of keyphrases is according to the independent clustering. |

Table 4: A summary of the baselines and variations against which our model is compared.

| Method | Restaurants | | |
|---|---|---|---|
| | Recall | Prec. | F-score |
| 1  Our model | 0.920 | 0.353 | **0.510** |
| 2  Random | 0.500 | 0.346 | 0.409 ∗ |
| 3  Keyphrase in text | 0.048 | 0.500 | 0.087 ∗ |
| 4  Keyphrase classifier | 0.769 | 0.353 | **0.484** ∗ |
| 5  Heuristic keyphrase classifier | 0.839 | 0.340 | **0.484** ∗ |
| 6  Model cluster in text | 0.227 | 0.385 | 0.286 ∗ |
| 7  Model cluster classifier | 0.721 | 0.402 | **0.516** |
| 8  Heuristic model cluster classifier | 0.731 | 0.366 | 0.488 ∗ |
| 9  Gold cluster model | 0.936 | 0.344 | **0.502** |
| 10  Gold cluster in text | 0.339 | 0.360 | 0.349 ∗ |
| 11  Gold cluster classifier | 0.693 | 0.366 | 0.479 ∗ |
| 12  Heuristic gold cluster classifier | 1.000 | 0.326 | 0.492 ◇ |
| 13  Independent cluster model | 0.745 | 0.363 | **0.488** ◇ |
| 14  Independent cluster in text | 0.220 | 0.340 | 0.266 ∗ |
| 15  Independent cluster classifier | 0.586 | 0.384 | 0.464 ∗ |
| 16  Heuristic independent cluster classifier | 0.592 | 0.386 | 0.468 ∗ |

Table 5: Comparison of the property predictions made by our model and a series of baselines and model variations in the restaurant domain, evaluated against expert semantic annotations. The results are divided according to experiment. The methods against which our model has significantly better results using approximate randomization are indicated with ∗ for $p \leq 0.05$, and ◇ for $p \leq 0.1$.

| Method | Restaurants | | | Cell Phones | | | Digital Cameras | | |
|---|---|---|---|---|---|---|---|---|---|
| | Recall | Prec. | F-score | Recall | Prec. | F-score | Recall | Prec. | F-score |
| 1 Our model | 0.923 | 0.623 | **0.744** | 0.971 | 0.537 | **0.692** | 0.905 | 0.586 | **0.711** |
| 2 Random | 0.500 | 0.500 | 0.500 * | 0.500 | 0.489 | 0.494 * | 0.500 | 0.501 | 0.500 * |
| 3 Keyphrase in text | 0.077 | 0.906 | 0.142 * | 0.171 | 0.529 | 0.259 * | 0.715 | 0.642 | 0.676 * |
| 4 Keyphrase classif. | 0.905 | 0.527 | **0.666** * | 1.000 | 0.500 | **0.667** | 0.942 | 0.540 | **0.687** ◇ |
| 5 Heur. keyphr. classif. | 0.997 | 0.497 | 0.664 * | 0.845 | 0.474 | 0.607 * | 0.845 | 0.531 | 0.652 * |
| 6 Model cluster in text | 0.416 | 0.613 | 0.496 * | 0.829 | 0.547 | 0.659 ◇ | 0.812 | 0.596 | 0.687 * |
| 7 Model cluster classif. | 0.859 | 0.711 | **0.778** † | 0.876 | 0.561 | **0.684** | 0.927 | 0.568 | 0.704 |
| 8 Heur. model classif. | 0.910 | 0.567 | 0.698 * | 1.000 | 0.464 | 0.634 ◇ | 0.942 | 0.568 | **0.709** |
| 9 Gold cluster model | 0.992 | 0.500 | 0.665 * | 0.924 | 0.561 | **0.698** | 0.962 | 0.510 | 0.667 * |
| 10 Gold cluster in text | 0.541 | 0.604 | 0.571 * | 0.914 | 0.497 | 0.644 * | 0.903 | 0.522 | 0.661 * |
| 11 Gold cluster classif. | 0.865 | 0.720 | **0.786** † | 0.810 | 0.559 | 0.661 | 0.874 | 0.674 | **0.761** |
| 12 Heur. gold classif. | 0.997 | 0.499 | 0.665 * | 0.969 | 0.468 | 0.631 ◇ | 0.971 | 0.508 | 0.667 * |
| 13 Indep. cluster model | 0.984 | 0.528 | 0.687 * | 0.838 | 0.564 | **0.674** | 0.945 | 0.519 | **0.670** * |
| 14 Indep. cluster in text | 0.382 | 0.569 | 0.457 * | 0.724 | 0.481 | 0.578 * | 0.469 | 0.476 | 0.473 * |
| 15 Indep. cluster classif. | 0.753 | 0.696 | **0.724** | 0.638 | 0.472 | 0.543 * | 0.496 | 0.588 | 0.538 * |
| 16 Heur. indep. classif. | 0.881 | 0.478 | 0.619 * | 1.000 | 0.464 | 0.634 ◇ | 0.969 | 0.501 | 0.660 * |

Table 6: Comparison of the property predictions made by our model and a series of baselines and model variations in three product domains, as evaluated against author free-text annotations. The results are divided according to experiment. The methods against which our model has significantly better results using approximate randomization are indicated with * for $p \leq 0.05$, and ◇ for $p \leq 0.1$. Methods which perform significantly better than our model with $p \leq 0.05$ are indicated with †.

$i$ appears $n_i$ times, then expected precision is $\sum_{i=1}^{m} \frac{n_i}{mn}$. For instance, for the restaurants gold standard evaluation, the six tested properties appeared a total of 249 times over 120 documents, yielding an expected precision of 0.346.

- *Keyphrase in text:* A keyphrase is supported by a document if it appears verbatim in the text. Precision should be high while recall will be low, because the model is unable to detect paraphrases of the keyphrase in the text. For instance, for the first review from Figure 1, "cleanliness" would be supported because it appears in the text; however, "healthy" would not be supported, even though the synonymous "great nutrition" does appear.

- *Keyphrase classifier:*[9] A separate discriminative classifier is trained for each keyphrase. Positive examples are documents that are labeled by the author with the keyphrase; all other documents are considered to be negative examples. Consequently, for any particular keyphrase, documents labeled with synonymous keyphrases would be among the negative examples. A keyphrase is supported by a document if that keyphrase's classifier returns a positive prediction.

  We use support vector machines, built using SVM$^{light}$ (Joachims, 1999) with the same features as our model, *i.e.,* word counts.[10] To partially circumvent the imbalanced positive/negative data problem, we tuned prediction thresholds on a development set to maximize F-score, in the same manner that we tuned thresholds for our model.

- *Heuristic keyphrase classifier:* This baseline is similar to *keyphrase classifier* above, but attempts to mitigate some of the noise inherent in the training data. Specifically, any given positive example document may contain text unrelated to the given keyphrase. We attempt to reduce this noise by removing from the positive examples all sentences that have no word overlap with the given keyphrase. A keyphrase is supported by a document if that keyphrase's classifier returns a positive prediction.[11]

Lines 2-5 of Tables 5 and 6 present these results, using both gold annotations and the original authors' annotations for testing. Our model outperforms these three baselines in all evaluations with strong statistical significance.

The *keyphrase in text* baseline fares poorly: its F-score is below the random baseline in three of the four evaluations. As expected, the recall of this baseline is usually low because it requires keyphrases to appear verbatim in the text. The precision is somewhat better, but the presence of a significant number of false positives indicates that the presence of a keyphrase in the text is not necessarily a reliable indicator of the associated semantic property.

Interestingly, one domain in which *keyphrase in text* does perform well is digital cameras. We believe that this is because of the prevalence of specific technical terms in the keyphrases used in this domain, such as "zoom" and "battery life." Such technical terms are also frequently used in the review text, making the recall of *keyphrase in text* substantially higher in this domain than in the other evaluations.

---

9. Note that the classifier results reported in the initial publication (Branavan, Chen, Eisenstein, & Barzilay, 2008) were obtained using the default parameters of a maximum entropy classifier. Tuning the classifier's parameters allowed us to significantly improve performance of all classifier baselines.

10. In general, SVMs have the additional advantage of being able to incorporate arbitrary features, but for the sake of comparison we restrict ourselves to using the same features across all methods.

11. We thank a reviewer for suggesting this baseline.

The *keyphrase classifier* baseline outperforms the *random* and *keyphrase in text* baselines, but still achieves consistently lower performance than our model in all four evaluations. Notably, the performance of *heuristic keyphrase classifier* is worse than *keyphrase classifier* except in one case. This alludes to the difficulty of removing the noise inherent in the document text.

Overall, these results indicate that methods which learn and predict keyphrases without accounting for their intrinsic hidden structure are insufficient for optimal property prediction. This leads us toward extending the present baselines with clustering information.

It is important to assess the consistency of the evaluation based on free-text annotations (Table 6) with the evaluation that uses expert annotations (Table 5). While the absolute scores on the expert annotations dataset are lower than the scores with free-text annotations, the ordering of performance between the various automatic methods is the same across the two evaluation scenarios. This consistency is maintained in the rest of our experiments as well, indicating that for the purpose of relative comparison between the different automatic methods, our method of evaluating with free-text annotations is a reasonable proxy for evaluation on expert-generated annotations.

**Comparison against Clustering-based Approaches** The previous section demonstrates that our model outperforms baselines that do not account for the paraphrase structure of keyphrases. We now ask whether it is possible to enhance the baselines' performance by augmenting them with the keyphrase clustering induced by our model. Specifically, we introduce three more systems, none of which are "true" baselines, since they all use information inferred by our model.

- *Model cluster in text:* A keyphrase is supported by a document if it or any of its paraphrases appears in the text. Paraphrasing is based on our model's clustering of the keyphrases. The use of paraphrasing information enhances recall at the potential cost of precision, depending on the quality of the clustering. For example, assuming "healthy" and "great nutrition" are clustered together, the presence of "healthy" in the text would also indicate support for "great nutrition," and vice versa.

- *Model cluster classifier:* A separate discriminative classifier is trained for each cluster of keyphrases. Positive examples are documents that are labeled by the author with any keyphrase from the cluster; all other documents are negative examples. All keyphrases of a cluster are supported by a document if that cluster's classifier returns a positive prediction. Keyphrase clustering is based on our model. As with *keyphrase classifier*, we use support vector machines trained on word count features, and we tune the prediction thresholds for each individual cluster on a development set.

    Another perspective on *model cluster classifier* is that it augments the simplistic text modeling portion of our model with a discriminative classifier. Discriminative training is often considered to be more powerful than equivalent generative approaches (McCallum et al., 2005), leading us to expect a high level of performance from this system.

- *Heuristic model cluster classifier:* This method is similar to *model cluster classifier* above, but with additional heuristics used to reduce the noise inherent in the training data. Positive example documents may contain text unrelated to the given cluster. To reduce this noise, sentences that have no word overlap with any of the cluster's keyphrases are removed. All keyphrases of a cluster are supported by a document if that cluster's classifier returns a positive prediction. Keyphrase clustering is based on our model.

Lines 6-8 of Tables 5 and 6 present results for these methods. As expected, using a clustering of keyphrases with the baseline methods substantially improves their recall, with low impact on precision. *Model cluster in text* invariably outperforms *keyphrase in text* — the recall of *keyphrase in text* is improved by the addition of clustering information, though precision is worse in some cases. This phenomenon holds even in the cameras domain, where *keyphrase in text* already performs well. However, our model still significantly outperforms *model cluster in text* in all evaluations.

Adding clustering information to the classifier baseline results in performance that is sometimes better than our model's. This result is not surprising, because *model cluster classifier* gains the benefit of our model's robust clustering while learning a more sophisticated classifier for assigning properties to texts. The resulting combined system is more complex than our model by itself, but has the potential to yield better performance. On the other hand, using a simple heuristic to reduce the noise present in the training data consistently hurts the performance of the classifier, possibly due to the reduction in the amount of training data.

Overall, the enhanced performance of these methods, in contrast to the keyphrase baselines, is aligned with previous observations in entailment research (Dagan, Glickman, & Magnini, 2006), confirming that paraphrasing information contributes greatly to improved performance in semantic inference tasks.

**The Impact of Paraphrasing Quality**   The previous section demonstrates one of the central claims of this paper: accounting for paraphrase structure yields substantial improvements in semantic inference when using noisy keyphrase annotations. A second key aspect of our research is the idea that clustering quality benefits from tying the clusters to hidden topics in the document text. We evaluate this claim by comparing our model's clustering against an independent clustering baseline. We also compare against a "gold standard" clustering produced by expert human annotators. To test the impact of these clustering methods, we substitute the model's inferred clustering with each alternative and examine how the resulting semantic inferences change. This comparison is performed for the semantic inference mechanism of our model, as well as for the *model cluster in text*, *model cluster classifier* and *heuristic model cluster classifier* baselines.

To add a "gold standard" clustering to our model, we replace the hidden variables that correspond to keyphrase clusters with observed values that are set according to the gold standard clustering.[12] The only parameters that are trained are those for modeling text. This model variation, *gold cluster model*, predicts properties using the same inference mechanism as the original model. The baseline variations *gold cluster in text*, *gold cluster classifier* and *heuristic gold cluster classifier* are likewise derived by substituting the automatically computed clustering with gold standard clusters.

An additional clustering is obtained using only the keyphrase similarity information. Specifically, we modify our original model so that it learns the keyphrase clustering in isolation from the text, and only then learns the property language models. In this framework, the keyphrase clustering is entirely independent of the review text, because the text modeling is learned with the keyphrase clustering fixed. We refer to this modification of the model as *independent cluster model*. Because our model treats the document text as a mixture of latent topics, this is reminiscent of models such as supervised latent Dirichlet allocation (sLDA; Blei & McAuliffe, 2008), with the labels acquired by performing a clustering across keyphrases as a preprocessing step. As in the previous experiment, we introduce three new baseline variations — *independent cluster in text*, *independent cluster classifier* and *heuristic independent cluster classifier*.

---

12. The gold standard clustering was created as part of the evaluation procedure described in Section 6.1.1.

Lines 9-16 of Tables 5 and 6 present the results of these experiments. The *gold cluster model* produces F-scores comparable to our original model, providing strong evidence that the clustering induced by our model is of sufficient quality for semantic inference. The application of the expert-generated clustering to the baselines (lines 10, 11 and 12) yields less consistent results, but overall this evaluation provides little reason to believe that performance would be substantially improved by obtaining a clustering that was closer to the gold standard.

The *independent cluster model* consistently reduces performance with respect to the full joint model, supporting our hypothesis that joint learning gives rise to better prediction. The independent clustering baselines, *independent cluster in text*, *independent cluster classifier* and *heuristic independent cluster classifier* (lines 14 to 16), are also worse than their counterparts that use the model clustering (lines 6 to 8). This observation leads us to conclude that while the expert-annotated clustering does not always improve results, the independent clustering always degrades them. This supports our view that joint learning of clustering and text models is an important prerequisite for better property prediction.

| Clustering | Restaurants | Cell Phones | Digital Cameras |
|---|---|---|---|
| Model clusters | **0.914** | **0.876** | **0.945** |
| Independent clusters | 0.892 | 0.759 | 0.921 |

Table 7: Rand Index scores of our model's clusters, learned from keyphrases and text jointly, compared against clusters learned only from keyphrase similarity. Evaluation of cluster quality is based on the gold standard clustering.

Another way of assessing the quality of each automatically-obtained keyphrase clustering is to quantify its similarity to the clustering produced by the expert annotators. For this purpose we use the *Rand Index* (Rand, 1971), a measure of cluster similarity. This measure varies from zero to one, with higher scores indicating greater similarity. Table 7 shows the Rand Index scores for our model's full joint clustering, as well as the clustering obtained from *independent cluster model*. In every domain, joint inference produces an overall clustering that improves upon the keyphrase-similarity-only approach. These scores again confirm that joint inference across keyphrases and document text produces a better clustering than considering features of the keyphrases alone.

## 6.2 Summarizing Multiple Reviews

Our last experiment examines the multi-document summarization capability of our system. We study our model's ability to aggregate properties across a set of reviews, compared to baselines that aggregate by directly using the free-text annotations.

### 6.2.1 DATA AND EVALUATION

We selected 50 restaurants, with five user-written reviews for each restaurant. Ten annotators were asked to annotate the reviews for five restaurants each, comprising 25 reviews per annotator. They used the same six salient properties and the same annotation guidelines as in the previous restaurant annotation experiment (see Section 3). In constructing the ground truth, we label properties that are supported in at least three of the five reviews.

| Method | Recall | Prec. | F-score |
|---|---|---|---|
| Our model | 0.905 | 0.325 | **0.478** |
| Keyphrase aggregation | 0.036 | 0.750 | 0.068 $*$ |
| Model cluster aggregation | 0.238 | 0.870 | 0.374 $*$ |
| Gold cluster aggregation | 0.226 | 0.826 | 0.355 $*$ |
| Indep. cluster aggregation | 0.214 | 0.720 | 0.330 $*$ |

Table 8: Comparison of the aggregated property predictions made by our model and a series of baselines that use free-text annotations. The methods against which our model has significantly better results using approximate randomization are indicated with $*$ for $p \leq 0.05$.

We make property predictions on the same set of reviews with our model and the baselines presented below. For the automatic methods, we register a prediction if the system judges the property to be supported on at least two of the five reviews.[13] The recall, precision, and F-score are computed over these aggregate predictions, against the six salient properties marked by annotators.

### 6.2.2 AGGREGATION APPROACHES

In this evaluation, we run the trained version of our model as described in Section 6.1.1. Note that keyphrases are not provided to our model, though they are provided to the baselines.

The most obvious baseline for summarizing multiple reviews would be to directly aggregate their free-text keyphrases. These annotations are presumably representative of the review's semantic properties, and unlike the review text, keyphrases can be matched directly with each other. Our first baseline applies this notion directly:

- *Keyphrase aggregation:* A keyphrase is supported for a restaurant if at least two out of its five reviews are annotated verbatim with that keyphrase.

This simple aggregation approach has the obvious downside of requiring very strict matching between independently authored reviews. For that reason, we consider extensions to this aggregation approach that allow for annotation paraphrasing:

- *Model cluster aggregation:* A keyphrase is supported for a restaurant if at least two out of its five reviews are annotated with that keyphrase or one of its paraphrases. Paraphrasing is according to our model's inferred clustering.

- *Gold cluster aggregation:* Same as *model cluster aggregation*, but using the expert-generated clustering for paraphrasing.

- *Independent cluster aggregation:* Same as *model cluster aggregation*, but using the clustering learned only from keyphrase similarity for paraphrasing.

---

13. When three corroborating reviews are required, the baseline systems produce very few positive predictions, leading to poor recall. Results for this setting are presented in Appendix B.

### 6.2.3 RESULTS

Table 8 compares the baselines against our model. Our model outperforms all of the annotation-based baselines, despite not having access to the keyphrase annotations. Notably, *keyphrase aggregation* performs very poorly, because it makes very few predictions, as a result of its requirement of exact keyphrase string match. As before, the inclusion of keyphrase clusters improves the performance of the baseline models. However, the incompleteness of the keyphrase annotations (see Section 3) explains why the recall scores are still low compared to our model. By incorporating document text, our model obtains dramatically improved recall, at the cost of reduced precision, ultimately yielding a significantly improved F-score.

These results demonstrate that review summarization benefits greatly from our joint model of the review text and keyphrases. Naïve approaches that consider only keyphrases yield inferior results, even when augmented with paraphrase information.

## 7. Conclusions and Future Work

In this paper, we have shown how free-text keyphrase annotations provided by novice users can be leveraged as a training set for document-level semantic inference. Free-text annotations have the potential to vastly expand the set of training data available to developers of semantic inference systems; however, as we have shown, they suffer from lack of consistency and completeness. We overcome these problems by inducing a hidden structure of semantic properties, which correspond both to clusters of keyphrases and hidden topics in the text. Our approach takes the form of a hierarchical Bayesian model, which addresses both the text and keyphrases jointly.

Our model is implemented in a system that successfully extracts semantic properties of unannotated restaurant, cell phone, and camera reviews, empirically validating our approach. Our experiments demonstrate the necessity of handling the paraphrase structure of free-text keyphrase annotations; moreover, they show that a better paraphrase structure is learned in a joint framework that also models the document text. Our approach outperforms competitive baselines for semantic property extraction from both single and multiple documents. It also permits aggregation across multiple keyphrases with different surface forms for multi-document summarization.

This work extends an actively growing literature on document topic modeling. Both topic modeling and paraphrasing posit a hidden layer that captures the relationship between disparate surface forms: in topic modeling, there is a set of latent distributions over lexical items, while paraphrasing is represented by a latent clustering over phrases. We show these two latent structures can be linked, resulting in increased robustness and semantic coherence.

We see several avenues of future work. First, our model draws substantial power from features that measure keyphrase similarity. This ability to use arbitrary similarity metrics is desirable; however, representing individual similarity scores as random variables is a compromise, as they are clearly not independent. We believe that this problem could be avoided by modeling the generation of the entire similarity matrix jointly.

A related approach would be to treat the similarity matrix across keyphrases as an indicator of covariance structure. In such a model, we would learn separate language models for each keyphrase, but keyphrases that are rated as highly similar would be constrained to induce similar language models. Such an approach might be possible in a Gaussian process framework (Rasmussen & Williams, 2006).

Currently the focus of our model is to identify the semantic properties expressed in a given document, which allows us to produce a summary of those properties. However, as mentioned in Section 3, human authors do not give equal importance to all properties when producing a summary of pros and cons. One possible extension of this work would be to explicitly model the likelihood of each topic being annotated in a document. We might then avoid the current post-processing step that uses property-specific thresholds to compute final predictions from the model output.

Finally, we have assumed that the semantic properties themselves are unstructured. In reality, properties are related in interesting ways. Trivially, in the domain of reviews it would be desirable to model antonyms explicitly, *e.g.*, no restaurant review should be simultaneously labeled as having good and bad food. Other relationships between properties, such as hierarchical structures, could also be considered. This suggests possible connections to the *correlated topic model* of Blei and Lafferty (2006).

## Bibliographic Note

Portions of this work were previously presented in a conference publication (Branavan et al., 2008). The current article extends this work in several ways, most notably: the development and evaluation of a multi-document review summarization system that uses semantic properties induced by our method (Section 6.2); a detailed analysis of the distributional properties of free-text annotations (Section 3); and an expansion of the evaluation to include an additional domain and sets of baselines not considered in the original paper (Section 6.1.1).

## Acknowledgments

The authors acknowledge the support of National Science Foundation (NSF) CAREER grant IIS-0448168, the Microsoft Research New Faculty Fellowship, the U.S. Office of Naval Research (ONR), Quanta Computer, and Nokia Corporation. Harr Chen is supported by the National Defense Science and Engineering and NSF Graduate Fellowships. Thanks to Michael Collins, Zoran Dzunic, Amir Globerson, Aria Haghighi, Dina Katabi, Kristian Kersting, Terry Koo, Yoong Keok Lee, Brian Milch, Tahira Naseem, Dan Roy, Christina Sauper, Benjamin Snyder, Luke Zettlemoyer, and the journal reviewers for helpful comments and suggestions. We also thank Marcia Davidson and members of the NLP group at MIT for help with expert annotations. Any opinions, findings, conclusions or recommendations expressed in this article are those of the authors, and do not necessarily reflect the views of NSF, Microsoft, ONR, Quanta, or Nokia.

## Appendix A. Development and Test Set Statistics

Table 9 lists the semantic properties for each domain and the number of documents that are used for evaluating each of these properties. As noted in Section 6.1.1, the gold standard evaluation is complete, testing every property with each document. Conversely, the free-text evaluations for each property only use documents that are annotated with the property or its antonym — this is why the number of documents differs for each semantic property.

| Domain | Property | Development documents | Test Documents |
|---|---|---|---|
| Restaurants (gold) | *All properties* | 50 | 120 |
| Restaurants | Good food<br>Bad food | 88 | 179 |
| | Good price<br>Bad price | 31 | 66 |
| | Good service<br>Bad service | 69 | 140 |
| Cell Phones | Good reception<br>Bad reception | 33 | 67 |
| | Good battery life<br>Poor battery life | 59 | 120 |
| | Good price<br>Bad price | 28 | 57 |
| Cameras | Small<br>Large | 84 | 168 |
| | Good price<br>Bad price | 56 | 113 |
| | Good battery life<br>Poor battery life | 51 | 102 |
| | Great zoom<br>Limited zoom | 34 | 69 |

Table 9: Breakdown by property for the development and test sets used for the evaluations in section 6.1.2.

## Appendix B. Additional Multiple Review Summarization Results

Table 10 lists results of the multi-document experiment, with a variation on the aggregation — we require each automatic method to predict a property for three of five reviews to predict that property for the product, rather than two as presented in Section 6.2. For the baseline systems, this change causes a precipitous drop in recall, leading to F-score results that are substantially worse than those presented in Section 6.2.3. In contrast, the F-score for our model is consistent across both evaluations.

| Method | Recall | Prec. | F-score |
|---|---|---|---|
| Our model | 0.726 | 0.365 | **0.486** |
| Keyphrase aggregation | 0.000 | 0.000 | 0.000 * |
| Model cluster aggregation | 0.024 | 1.000 | 0.047 * |
| Gold cluster aggregation | 0.036 | 1.000 | 0.068 * |
| Indep. cluster aggregation | 0.036 | 1.000 | 0.068 * |

Table 10: Comparison of the aggregated property predictions made by our model and a series of baselines that only use free-text annotations. Aggregation requires three of five reviews to predict a property, rather than two as in Section 6.2. The methods against which our model has significantly better results using approximate randomization are indicated with $*$ for $p \leq 0.05$.

## Appendix C. Hyperparameter Settings

Table 11 lists the values of hyperparameters $\theta_0$, $\psi_0$, and $\phi_0$ used in all experiments for each domain. These values were arrived at through tuning on the development set. In all cases, $\lambda_0$ was set to $(1, 1)$, making Beta$(\lambda_0)$ the uniform distribution.

| Hyperparameters | Restaurants | Cell Phones | Cameras |
|---|---|---|---|
| $\theta_0$ | 0.0001 | 0.0001 | 0.0001 |
| $\psi_0$ | 0.001 | 0.0001 | 0.1 |
| $\phi_0$ | 0.001 | 0.0001 | 0.001 |

Table 11: Values of the hyperparameters used for each domain across all experiments.